\documentclass[10pt,twocolumn,letterpaper]{article}

\usepackage{cvpr}              
\usepackage[accsupp]{axessibility}
\usepackage{graphicx}
\usepackage{amsmath}
\usepackage{amsthm}
\usepackage{amssymb}
\usepackage{booktabs}
\usepackage{appendix}
\usepackage{multirow}
\usepackage{graphicx}
\usepackage{makecell}
\usepackage{microtype}
\usepackage[american]{babel}
\usepackage[accsupp]{axessibility}  
\usepackage[linesnumbered,ruled,vlined]{algorithm2e}
\newtheorem{theorem}{Theorem}
\newtheorem{lemma}{Lemma} 

\usepackage[dvipsnames]{xcolor}

\definecolor{cvprblue}{rgb}{0.21,0.49,0.74}
\usepackage[pagebackref,breaklinks,colorlinks,citecolor=cvprblue]{hyperref}

\def\confName{CVPR}
\def\confYear{2024}

\title{\textsc{Free}: Faster and Better Data-Free Meta-Learning}

\author{
Yongxian Wei\textsuperscript{\rm 1}
\quad 
Zixuan Hu\textsuperscript{\rm 1}
\quad
Zhenyi Wang\textsuperscript{\rm 2}
\quad 
Li Shen\textsuperscript{\rm 3,}\thanks{Corresponding authors: Li Shen and Chun Yuan} 
\quad 
Chun Yuan\textsuperscript{\rm 1,*}
\quad 
Dacheng Tao\textsuperscript{\rm 4}
\\
\textsuperscript{\rm 1}Tsinghua Shenzhen International Graduate School; 
\textsuperscript{\rm 2}University of Maryland, College Park\\ 
\textsuperscript{\rm 3}JD Explore Academy;
\textsuperscript{\rm 4}Nanyang Technological University\\
{\tt\small weiyx23@mails.tsinghua.edu.cn; huzixuan21@mails.tsinghua.edu.cn; zwang169@umd.edu}\\
{\tt\small mathshenli@gmail.com; yuanc@sz.tsinghua.edu.cn; dacheng.tao@ntu.edu.sg
}
}

\SetCommentSty{myCommentStyle}

\begin{document}
\maketitle
\begin{abstract}
Data-Free Meta-Learning (DFML) aims to extract knowledge from a collection of pre-trained models without requiring the original data, presenting practical benefits in contexts constrained by data privacy concerns. Current DFML methods primarily focus on the data recovery from these pre-trained models. However, they suffer from slow recovery speed and overlook gaps inherent in heterogeneous pre-trained models. In response to these challenges, we introduce the \textbf{F}aster and Bette\textbf{r} Data-Fr\textbf{e}e M\textbf{e}ta-Learning (\textsc{Free}) framework, which contains: (i) a meta-generator for rapidly recovering training tasks from pre-trained models; and (ii) a meta-learner for generalizing to new unseen tasks. Specifically, within the module Faster Inversion via Meta-Generator, each pre-trained model is perceived as a distinct task. The meta-generator can rapidly adapt to a specific task in just five steps, significantly accelerating the data recovery. Furthermore, we propose Better Generalization via Meta-Learner and introduce an implicit gradient alignment algorithm to optimize the meta-learner. This is achieved as aligned gradient directions alleviate potential conflicts among tasks from heterogeneous pre-trained models. Empirical experiments on multiple benchmarks affirm the superiority of our approach, marking a notable speed-up (20$\times$) and performance enhancement (1.42\% $\sim$ 4.78\%) in comparison to the state-of-the-art. Code is available \href{https://github.com/WalkerWorldPeace/FREE}{here}.
\end{abstract}    
\section{Introduction}
\label{sec:intro}
Data-Free Meta-Learning (DFML)~\cite{kwon2020repurposing,wang2022meta,hu2023architecture,hu2023learning} aims to derive knowledge from a collection of pre-trained models without necessitating the original data, enabling the adaptation of knowledge to new unseen tasks. Traditional meta-learning methods assume access to a collection of tasks with available training and testing data. However, in many real situations, such data is often unavailable~\cite{chen2019data,truong2021data,zheng2023learn,li2023deep}, primarily due to data privacy concerns, security risks, or usage rights. For example, numerous individuals and institutions release task-specific pre-trained models from diverse domains on platforms like GitHub or Hugging Face without training data released. Such real-world situations highlight the value of DFML: collect some pre-trained models with weaker generalization abilities, which likely originate from diverse domains online, and train a meta-learner with superior generalization ability for new tasks.

\begin{figure}[t]
  \centering
    \includegraphics[width=\linewidth]{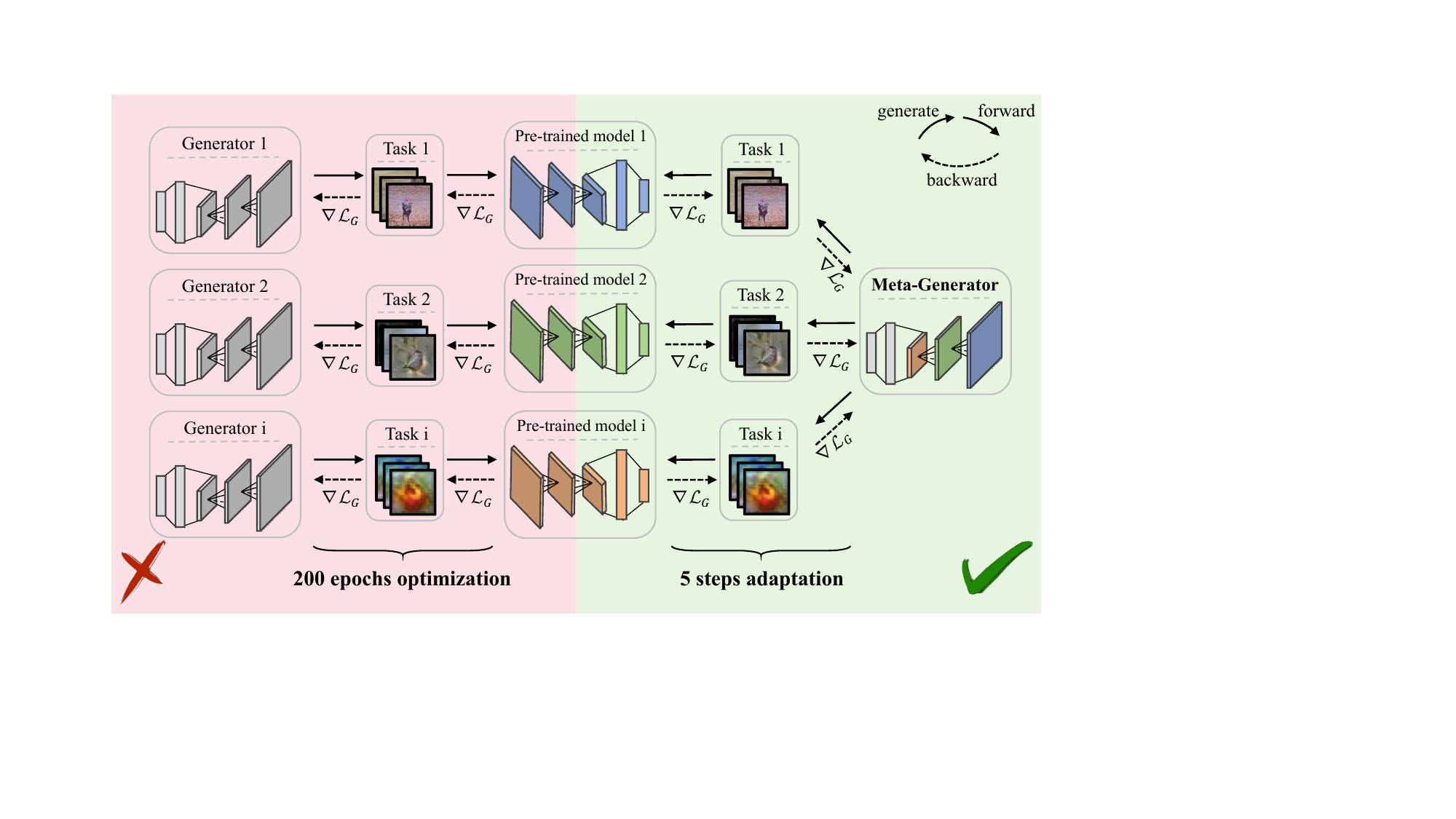}
    \vspace{-1em}
   \caption{Faster Inversion via Meta-Generator significantly enhances the efficiency of task generation. Tasks recovered from pre-trained models are used for training in the data-free setting. For each task, prior works need to train a specific generator with hundreds of generate-forward-backward iterations, while we only need a 5-step adaptation using the single meta-generator.}
   \label{fig:faster}
   \vspace{-1em}
\end{figure}

\begin{figure}[t]
\begin{minipage}[t]{\linewidth}
        \centering
       {\includegraphics[width=0.48\columnwidth]{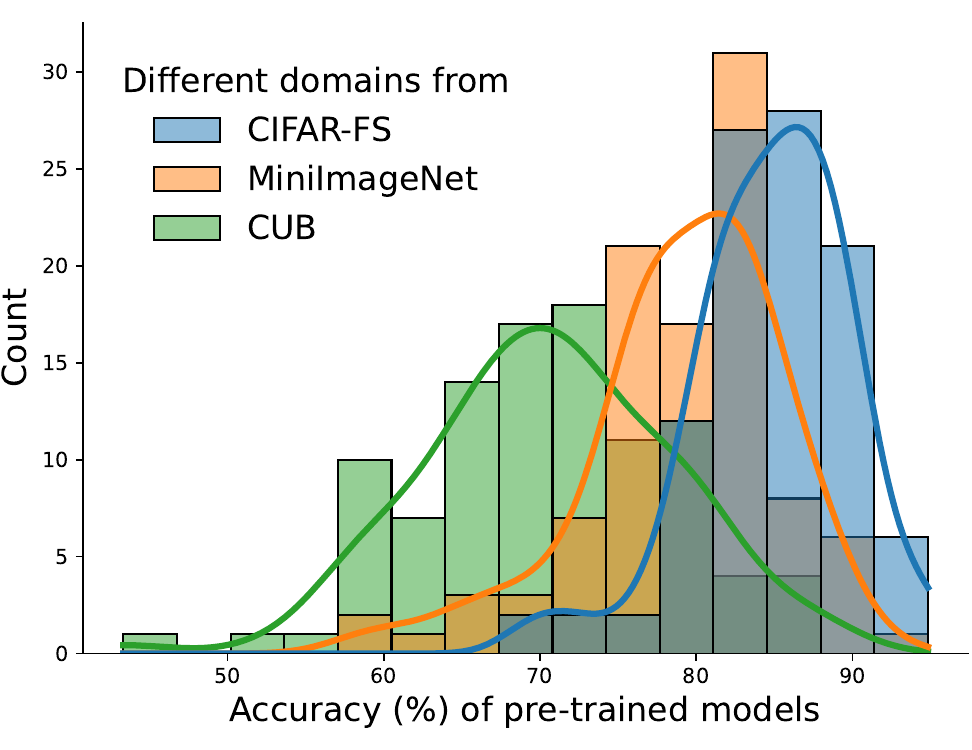} }
        {\includegraphics[width=0.49\columnwidth]{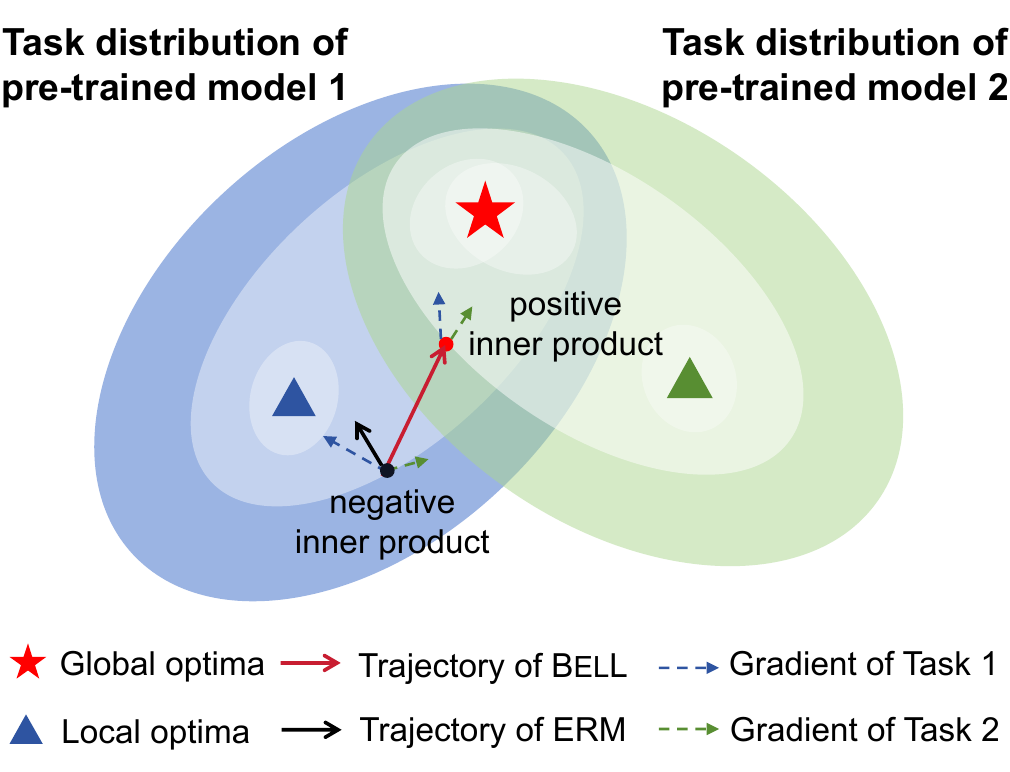} }
        \vspace{-1em}
         \caption{Pre-trained models from different domains inherently exhibit distribution differences. Pre-trained models, even when trained on different classes of the same dataset, display variations in performance quality. As a result, their recovered tasks naturally present a gap in the distribution. Overlooking such model heterogeneity will cause the meta-learner to bias towards specific tasks, leading to local optima. Our proposed \textsc{BelL} optimizes the meta-learner by encouraging a positive inner product of gradients across tasks, thus enhancing its generalization ability.}
	\label{fig:better}
\end{minipage}
\vspace{-1em}
\end{figure}

Existing DFML methods predominantly center on data recovery from pre-trained models. PURER~\cite{hu2023architecture} optimizes a learnable dataset through model inversion for each pre-trained model, subsequently sampling pseudo tasks for meta-learning. BiDf-MKD~\cite{hu2023learning}, on the other hand, trains generators using multiple black-box APIs, generating both the support and query sets distinctly for meta-learning.

In this work, we rethink DFML from two perspectives detailed in \cref{sec:rethink}. Through this re-evaluation, we find two pivotal limitations in existing methods: (i) \textbf{time-consuming data recovery processes}, and (ii) \textbf{overlooking the heterogeneity among different pre-trained models}. Specifically, prior research emphasizes generator learning \emph{at the instance level}, where each pre-trained model is paired with a unique generator, designed to generate images consistent with specific training distribution. Training such a generator often necessitates hundreds of generate-forward-backward iterations, making the recovery process of training data considerably time-consuming. Moreover, due to the varying model architectures and qualities of different pre-trained models, some of which may have relatively low accuracy, and the possibility that they originate from distinct domains (see \cref{fig:better}), their recovered tasks inherently present a distribution gap. We also present a $t$-SNE visualization to show the gap among them in \cref{fig:domain}. Applying simple Empirical Risk Minimization (ERM) might lead the meta-learner to be biased towards specific tasks, compromising its performance on others and limiting its generalization to unseen tasks.
Armed with these insights, we propose \textbf{F}aster and Bette\textbf{r} Data-Fr\textbf{e}e M\textbf{e}ta-Learning (\textsc{Free}), a unified framework that contains a meta-generator and a meta-learner. The meta-generator rapidly recovers specific tasks from pre-trained models for training the meta-learner, significantly accelerating the data recovery process. Concurrently, the meta-learner alleviates the gap among different tasks recovered from heterogeneous models, enhancing generalization to new unseen tasks.

For \textbf{f}aster \textbf{i}nversion \textbf{v}ia m\textbf{e}ta-generator (\textsc{FIve}), we introduce the concept of a meta-generator learned \emph{at the task level} (see \cref{fig:faster}). The meta-generator is trained across all pre-trained models for capturing shared representational knowledge. Drawing inspiration from meta-learning, we interpret \emph{each pre-trained model as a distinct task}. The meta-generator is designed to yield minimal loss for each pre-trained model, adapting through only five steps using its self-generated data. The inner loop's objective is to hone the rapid adaptability of the meta-generator, while the outer loop focuses on its generalization across multiple pre-trained models.
For \textbf{be}tter genera\textbf{l}ization via meta-\textbf{l}earner (\textsc{BelL}), we introduce an implicit gradient alignment algorithm across tasks from different pre-trained models, utilizing multi-task knowledge distillation to train the meta-learner. We treat the current task as the outer loop and its conflicting tasks as the inner loop, encouraging the optimization path to be suitable for all tasks. For instance, if the gradient directions of $\boldsymbol{g}_i$ and $\boldsymbol{g}_j$ are in alignment such that $\boldsymbol{g}_i\cdot \boldsymbol{g}_j>0$, gradient descent updates will alleviate conflicts between different tasks~\cite{shi2021gradient}. By advocating a shared gradient direction, our meta-learner can extract common features across tasks/domains. This ensures that the meta-learner excels even on new unseen tasks, achieving superior generalization compared to ERM.

We conduct comprehensive experiments on three meta-learning benchmark datasets, \ie, \emph{mini-ImageNet}, \emph{CIFAR-FS}, and \emph{CUB}, for demonstrating our superiority over existing DFML methods. Compared to the state-of-the-art, our approach not only achieves a 20$\times$ speed-up but also shows an improvement (+ 1.42\% $\sim$ 4.78\%). Furthermore, our approach effectively tackles the model heterogeneity in challenging multi-domain and multi-architecture scenarios.

In summary, our main contributions are three-fold:
\begin{itemize}
\item For the first time, we have a closer look at DFML, highlighting the pressing importance of addressing efficiency dilemma and model heterogeneity.
\item To accelerate data recovery processes, we innovatively treat pre-trained models as tasks, focusing on a rapidly adaptive meta-generator. Recognizing the heterogeneity among different pre-trained models, we incorporate gradient alignment into the DFML framework. This alleviates inherent conflicts/gaps across tasks/domains, thereby improving the meta-learner's generalization.
\item Experiments on various benchmarks demonstrate the superiority of the proposed approach. Furthermore, we provide comprehensive discussions to elucidate the working mechanism of each component in our approach.
\end{itemize}

\section{Related Work}
\label{sec:related}

\noindent
\textbf{Meta-learning}~\cite{snell2017prototypical,rajeswaran2019meta,jamal2019task,rajasegaran2020itaml,yao2021meta,chen2021meta,chavan2022dynamic,Qin2023CVPR,wei2024task}, aka ``learning-to-learn'', aims to acquire general prior knowledge from a large collection of tasks, enabling learning systems to rapidly adapt to novel categories using only a few examples. Further, \citet{nichol2018first} introduce Reptile as an evolution of MAML~\cite{finn2017model}. This work underscores the link between the inner loop update and the maximization of the gradient inner product. Reptile aims to align gradients across minibatches from the same task, which enhances within-task generalization. In contrast, our approach formulates the inner loop using multiple tasks recovered from heterogeneous pre-trained models, which fosters across-task generalization.

For the heterogeneity in meta-learning, Yao et~al. propose to cluster tasks into hierarchical structures enabling customized knowledge to heterogeneous tasks~\cite{yao2019hierarchically}, or disentangle the meta-learner as a graph with different knowledge blocks to learn from heterogeneous tasks~\cite{yao2020online}.

\noindent
\textbf{Data-free meta-learning (DFML)}~\cite{kwon2020repurposing,hu2023architecture,hu2023learning,weitask,wei2024meta} is to ensure adaptation of acquired knowledge to new unseen tasks without training data. By leveraging multiple pre-trained models with weaker generalization abilities, which likely originate from diverse domains online, the intent is to learn a meta-learner with superior generalization ability. Regarding previous works, \citet{wang2022meta} suggest predicting the meta initialization using a neural network that operates in the parameter space. Recently, PURER~\cite{hu2023architecture} introduces a progressive method to synthesize a series of pseudo-tasks, inversing training data from each pre-trained model. BiDf-MKD~\cite{hu2023learning} meta-learns the meta initialization by transferring meta knowledge from a collection of black-box APIs via zero-order gradient estimation. \citet{fang2022up} propose to reuse common features across different images, rather than optimizing each image independently.
In our study, we apply meta-learning principles at the pre-trained model level to accelerate the training process.

\noindent
\textbf{Gradient alignment}~\cite{shi2021gradient,eshratifar2018gradient,lee2022contextual} emphasizes the alignment or consistency of gradients during model training, and empirical evidence has shown its advantages in fields such as continual learning~\cite{wang2022learning}, federated learning~\cite{dandi2022implicit}, and multi-task learning~\cite{guangyuan2022recon}. Specifically, \citet{ren2018learning} focus on employing the gradient inner product to assign importance weights to training examples. PCGrad~\cite{yu2020gradient} introduces a gradient surgery technique that projects the gradient of one task onto the normal plane of the conflicting gradient of another task. Recently, \citet{patel2023learning} identify an implicit aligning factor between the knowledge-retention and knowledge-acquisition in data-free knowledge distillation. Nonetheless, we identify and address the heterogeneity that exists among different pre-trained models, introducing a more effective task-aligned meta-learning approach.
\section{A Closer Look at DFML}
In this section, we first review how DFML learns from a collection of pre-trained models. Then, we rethink DFML from two perspectives: efficiency and model heterogeneity.
\subsection{Preliminary}

\noindent
\textbf{DFML Setup.}
We are given only a collection of pre-trained models $\mathcal{M}_{pool}=\{M_i\}$, each designed to solve a specific task without accessing their training data. DFML aims to learn a meta-learner $F(\cdot;\boldsymbol{\theta})$ that can be rapidly transferred to new unseen tasks using pre-trained models $\mathcal{M}_{pool}$. During meta-testing, we sample ``$N$-way $K$-shot'' tasks. For such a task, there are $N$ classes and each class only has $K$ labeled samples $(\boldsymbol{S}, \boldsymbol{Y_S})$ named the support set, and $U$ unlabelled samples per class $\boldsymbol{Q}$ called the query set, where $\boldsymbol{Y_Q}$ is the corresponding ground truth. We use the support set $(\boldsymbol{S}, \boldsymbol{Y_S})$ to adapt the meta-learner $F(\cdot;\boldsymbol{\theta})$ to each specific task. The query set $\boldsymbol{Q}$ is what we actually need to predict.

\noindent
\textbf{Model inversion.}
Model inversion~\cite{frikha2023towards,patel2023learning,yu2023data,liu2021data} aims to recover the training data $\boldsymbol{\hat{X}}$ from the pre-trained model $M$ as an alternative to the inaccessible original data $\boldsymbol{X}$.
So a generator $G(\cdot;\boldsymbol{\theta}_G)$ can be introduced to estimate the distribution of $\boldsymbol{X}$ by designing an inversion loss with the pre-trained model $M$. Specifically, given a latent code $\boldsymbol{z}$ as a low-dimensional representation, the generator maps $\boldsymbol{z}$ to the intended data approximation $\boldsymbol{\hat{x}}$.
In this case, the generator is only responsible for a small part of the data distribution. Compared with the pixel updating strategy used in \cite{hu2023architecture} that updates different pixels independently, the generator can provide stronger regularization on pixels because they are produced from the shared weights.

\subsection{Rethinking Existing DFML Methods}\label{sec:rethink}

\noindent
\textbf{Efficiency dilemma.}
Model inversion generally requires a \emph{generate-forward-backward} computation for each optimization step, with the gradient back-propagation process being the most time-intensive~\cite{gruslys2016memory,sanyal2022towards}. As the structure of the pre-trained model expands, or the scale of the generated data increases, the time and computational costs rise substantially. We quantitatively measure the GFLOPs required for the gradient back-propagation of a single image through different model architectures, as shown in \cref{table:flops}. The total computation for inversion can be expressed as $N_{iter} \times N_{img} \times$~GFLOPs.
Prior works optimize pixels individually or train a unique generator from scratch for each pre-trained model to reconstruct specific training distribution. Such pixel optimization or generator training often demands hundreds of generate-forward-backward computations. However, certain foundational features and textures are \emph{shared} across different tasks, typically as the shallow layers of the generator~\cite{raghu2019rapid}. If we train a meta-generator that retains common initialization parameters, rapid adaptation to specific tasks can be achieved in a few steps.
\begin{table}[tb]
\footnotesize
\centering
\caption{\small GFLOPs of one image's gradient back-propagation.}
\vspace{-0.5em}
\resizebox{0.8\linewidth}{!}{
\begin{tabular}{cccc}
\toprule
\bf{Resolution} & Conv-4     & ResNet-18     & ResNet-50     \\\midrule
32 $\times$ 32	 & 0.03 & 1.12 & 2.62 \\\midrule
84 $\times$ 84 & 0.23 & 7.87 &  18.36\\\midrule
224	$\times$ 224 & 1.63 & 54.67 & 128.34\\
\bottomrule                
\end{tabular}}
\vspace{-1em}
\label{table:flops}
\end{table}

\begin{figure*}[t]
  \centering
    \includegraphics[width=\linewidth]{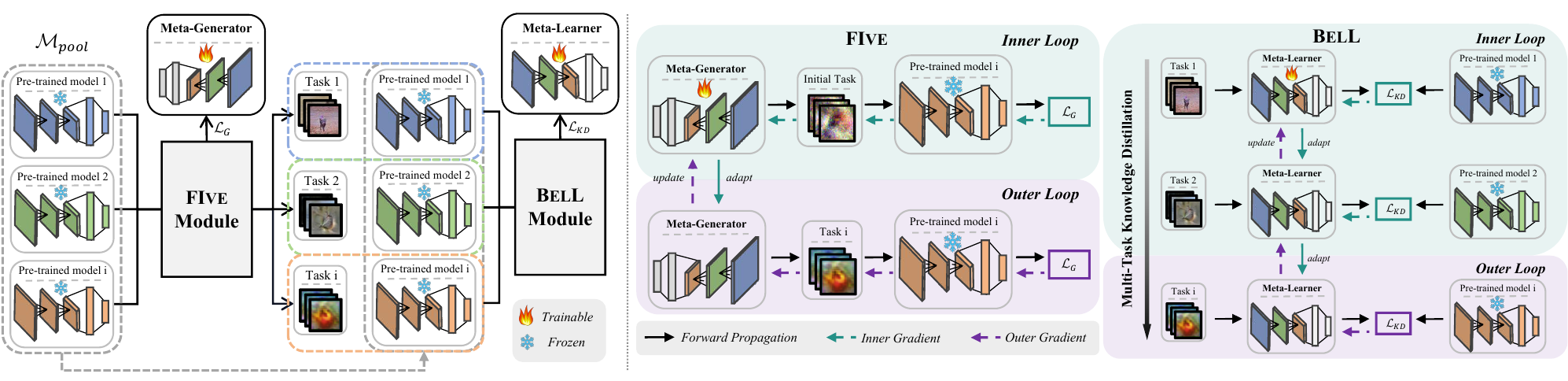}
    \vspace{-1em}
   \caption{An illustration of the proposed DFML framework. The framework consists of multiple pre-trained models ($\mathcal{M}_{pool}$), a meta-generator and a meta-learner. The model inversion loss ($\mathcal{L}_G$) optimizes the meta-generator, while the knowledge distillation loss ($\mathcal{L}_{KD}$) optimizes the meta-learner. After adapting to pre-trained models, the meta-generator recovers specific tasks. The meta-learner learns from recovered tasks and their respective pre-trained models by multi-task knowledge distillation.}
   \label{fig:method}
   \vspace{-1em}
\end{figure*}

\noindent
\textbf{Model heterogeneity.}
Pre-trained models which contain random classes can be perceived as sub-domains with their diverse data distributions, due to statistical properties differences such as pixel intensity or texture variation~\cite{chen2021variational,song2020robust}. Diverse model architectures also contribute to model heterogeneity, influencing how data is represented. Moreover, our knowledge regarding the quality of pre-trained models is limited, as we have \emph{no data} for evaluation. Given the chance that they stem from different domains, the tasks recovered from them inherently exhibit a distribution gap. We provide a $t$-SNE visualization to illustrate the gap among them in \cref{fig:domain}. Since neural networks tend to favor shortcuts and frequently learn simpler biases~\cite{geirhos2020shortcut}, a straightforward ERM across different tasks will cause the meta-learner to develop an undue bias towards specific tasks, limiting its generalization capabilities for more challenging unseen tasks. As shown in \cref{tab:ablation}, applying ERM leads to a 3.04\% decrease in our approach's performance on the unseen test set.
\section{Methodology}
Inspired by the above observation, we propose a unified framework \textsc{Free} as illustrated in \cref{fig:method}. In \cref{method:1}, we describe our proposed \textsc{FIve} module for accelerating model inversion, followed by adaptive task recovery and meta-generator learning. In \cref{method:2}, we propose \textsc{BelL} module to align gradients of different tasks recovered from the heterogeneous model, training a meta-learner for new tasks.

\vspace{-0.1em}
\subsection{Faster Inversion via Meta-Generator (\textsc{FIve})}\label{method:1}
Synthesizing training data from pre-trained models is a primary task in DFML. We propose viewing these pre-trained models as distinct \emph{tasks} and employing the meta-learning inspired strategy to train a \emph{meta-generator}. This meta-generator can rapidly adapt to a given pre-trained model within $k$ steps, generating task-specific data for training the meta-learner, significantly reducing the time for model inversion. We initially discuss how to inverse data from pre-trained models using the meta-generator, followed by how to learn the meta-generator's parameters.

\noindent
\textbf{Adaptive Task Recovery.}
The meta-generator $G(\cdot;\boldsymbol{\theta}_G)$ takes the standard Gaussian noise $\boldsymbol{Z}$ as inputs and outputs the recovered data $\boldsymbol{\hat{X}} = G(\boldsymbol{Z} ; \boldsymbol{\theta}_G)$.
For a pre-trained model $M$, we aim to recover a pseudo task $\mathcal{T}=\{(\boldsymbol{\hat{X}},\boldsymbol{Y})\}$ via the meta-generator. In order to obtain the training distribution of the pre-trained model, we need to perform $k$ steps of rapid adaptation on the meta-generator. We utilize the DeepInversion loss~\cite{yin2020dreaming} for updating $\boldsymbol{Z}$ and $\boldsymbol{\theta}_G$, which contains a classification loss to condition the recovered data on pre-defined target label $\boldsymbol{Y}$, and an adversarial loss to encourage data diversity by maximizing the KL divergence between the pre-trained model $M$ and the meta-learner $F(\cdot;\boldsymbol{\theta})$:
\begin{equation}\label{eq:1}
\underset{\boldsymbol{Z},\boldsymbol{\theta}_{G}}{\mathcal{L}_{G}}= {CE}(M(\hat{\boldsymbol{X}}),\boldsymbol{Y})-\eta\cdot{KL}(M(\hat{\boldsymbol{X}}),F(\hat{\boldsymbol{X}};\boldsymbol{\theta})),
\end{equation}
where $\eta=\mathbb{I}\{\mathop{\arg\max}M(\hat{\boldsymbol{X}})=\mathop{\arg\max}F(\hat{\boldsymbol{X}};\boldsymbol{\theta})\}$ and $\hat{\boldsymbol{X}}=G(\boldsymbol{Z}; \boldsymbol{\theta}_G)$. The function $\mathbb{I}(\cdot)$ is an indicator to enable $\hat{\boldsymbol{X}}$ with the same prediction from the pre-trained model and the meta-learner ($\eta$ = 1), otherwise disable it ($\eta$ = 0).

For each pre-trained model $M_i$, we first clone the meta-generator $G(\cdot;\boldsymbol{\theta}_G)$ and randomly sample a latent code $\boldsymbol{Z}$. After a $k$-step adaptation as the inner loop, we obtain task-specific parameters $(\boldsymbol{Z}^i,\boldsymbol{\theta}_{G}^{i}) $. Then, we can recover specific data $\boldsymbol{\hat{X}}_i = G(\boldsymbol{Z}^{i} ; \boldsymbol{\theta}^{i}_G)$ from the pre-trained model $M_i$.

\noindent
\textbf{Meta-Generator Learning.}
In the inner loop, the adaptation process provides a sequence of losses $\mathcal{L}_G^0,\mathcal{L}_G^1,\ldots,\mathcal{L}_G^{k-1}$, \ie, losses on different data generated by the updating generator over $k$ steps. To allow the meta-generator to build an internal representation suitable for a wide range of pre-trained models, the outer loop attempts to make the task-specific parameters reachable within the $k$-step adaptation:
\begin{align}
\label{eq:2}
&\min_{\boldsymbol{\theta}_G}\mathcal{L}_G^k(\boldsymbol{\hat{X}}_i)=\mathcal{L}_G^k(G(\boldsymbol{Z}^i ; \boldsymbol{\theta}_{G}^{i})),\nonumber\\
\mathrm{s.t.}(\boldsymbol{Z}^i,\boldsymbol{\theta}_{G}^{i})&=(\boldsymbol{Z},\boldsymbol{\theta}_{G})-\nabla\mathcal{L}_G^0-\nabla\mathcal{L}_G^1-\ldots-\nabla\mathcal{L}_G^{k-1}.
\end{align}

When optimizing \cref{eq:2} via gradient descent, we also accelerate the meta-generator learning in back-propagation. We compute the gradient at $\boldsymbol{\theta}_{G}^{i}$ alternatively~\cite{finn2017model}, and use the approximated gradient $\nabla_{\boldsymbol{\theta}_{G}^{i}}\mathcal{L}_{G}^k$ to update the meta-generator $G(\cdot;\boldsymbol{\theta}_G)$ as the outer loop (\cf \cref{alg:alg1}).

\begin{algorithm}[tb]
\small
\DontPrintSemicolon
\SetKwInOut{Input}{Input}\SetKwInOut{Output}{Output}\SetKwInOut{Require}{Require}
\textbf{Input: }{Meta-generator $G(\cdot;\boldsymbol{\theta}_G)$; meta-learner $F(\cdot;\boldsymbol{\theta})$; pre-trained models $\{M_i\}_{i=1}^b$ in a batch.
}\\
\textbf{Output: }{Optimized meta-generator $G(\cdot;\boldsymbol{\theta}_G)$;\,\,\,\,\,\,\,\,\, 
recovered tasks $\{\mathcal{T}_{i}\}_{i=1}^b$ in a batch.}\\

\For{ {\rm index} $i \leftarrow 1$ \KwTo $b$}{
Randomly sample $\boldsymbol{Z}$ and clone $G(\cdot;\boldsymbol{\theta}_G)$ \\
\tcp{After a $k$-step adaptation}
Obtain task-specific $(\boldsymbol{Z}^i,\boldsymbol{\theta}_{G}^{i})$ from $M_i$ \wrt \cref{eq:1}\\
Generate recovered data $\boldsymbol{\hat{X}}_i = G(\boldsymbol{Z}^{i} ; \boldsymbol{\theta}^{i}_G)$\\
\tcp{Meta-generator learning}
Compute $\boldsymbol{g}_G^i = \nabla_{\boldsymbol{\theta}_{G}^{i}}\mathcal{L}_{G}^k(\boldsymbol{\hat{X}}_{i})$ \wrt \cref{eq:2}\\
}
\tcp{Outer loop for meta-generator}
$\text{Update
} \boldsymbol{\theta}_G\leftarrow\boldsymbol{\theta}_G-\gamma \frac1b\sum_{i=1}^b\boldsymbol{g}_G^i$
\caption{ \textsc{FIve}}
\label{alg:alg1}
\end{algorithm}

\subsection{Better Generalization via Meta-Learner (\textsc{BelL})}\label{method:2}
Pseudo tasks recovered from heterogeneous pre-trained models exhibit the distribution gap. To utilize them for jointly training a meta-learner, we propose to align gradient directions of different tasks, facilitated by a multi-task knowledge distillation meta-learning algorithm.

\noindent
\textbf{Multi-Task Knowledge Distillation.}
Upon obtaining the recovered tasks from pre-trained models, we propose executing knowledge distillation to transfer the task-specific knowledge from the pre-trained model (acting as the teacher) to the meta-learner (acting as the student) using the recovered task. We optimize the meta-learner by minimizing the disagreement of predictions between them. This approach is favored because knowledge distillation offers richer supervision. It leverages the semantic class relationships present in the soft-label predictions from the pre-trained model $M$, as opposed to solely relying on the hard-label supervision from the generated data $\boldsymbol{\hat{X}}$.

For acquiring general knowledge from task-specific pre-trained models, we learn from multiple recovered tasks sampled from the pre-trained model pool $\mathcal{M}_{pool}$ at the same time. A challenge arises when these tasks exhibit conflicting optimization directions (their gradients inner product $<0$). In such scenario, optimizing for task $\mathcal{T}_i$ could inadvertently deteriorate the performance on task $\mathcal{T}_j$ and vice versa. Unlike conventional meta-learning approaches which split a single task into a support set for the inner loop and a query set for the outer loop, we seek to align conflicting tasks by learning different tasks for the inner and outer loops. Specifically, we optimize the meta-learner $F(\cdot;\boldsymbol{\theta})$ across recovered tasks in a sequence-by-sequence manner. In each sequence, the meta-learner is optimized for the task $\mathcal{T}_i$, while minimizing interference with other tasks:
\begin{align}
\min_{\boldsymbol{\theta}}\mathbb{E}_{\mathcal{T}_{i}\sim \mathcal{P}_{M}}&\mathcal{L}_{KD}(\mathcal{T}_{i};\boldsymbol{\widetilde{\theta}}) \triangleq {KL}(M_{i}(\hat{\boldsymbol{X}}_{i}),F(\hat{\boldsymbol{X}}_{i};\boldsymbol{\widetilde{\theta}})),\nonumber\\
\mathrm{s.t.~}&\boldsymbol{\widetilde{\theta}} = \min_{\boldsymbol{\theta}}\mathbb{E}_{\mathcal{T}_{j}\sim\mathcal{I}(\mathcal{T}_{i})}\mathcal{L}_{KD}(\mathcal{T}_{j};\boldsymbol{\theta}), \label{eq:3}
\end{align}
where $\mathcal{P}_{M}$ represents the joint task distribution of pre-trained models $\mathcal{M}_{pool}$, and $\mathcal{I}(\mathcal{T}_{i})$ denotes the set of other tasks in the sequence that come before task $\mathcal{T}_{i}$.

\noindent
\textbf{Implicit Gradient Alignment.}
To understand how it results in the desired alignment between different tasks, we can analyze the proposed objective in \cref{eq:3}. We employ Taylor's expansion to elucidate the connection between the gradient $\nabla_{\boldsymbol{\widetilde{\theta}}}\mathcal{L}_{KD}(\mathcal{T}_{i};\boldsymbol{\widetilde{\theta}})$ and gradient at initial point $\nabla_{\boldsymbol{\theta}}\mathcal{L}_{KD}(\mathcal{T}_{i};\boldsymbol{\theta})$, as described in \cref{lemma1}.

\begin{lemma}\label{lemma1}
If $\mathcal{L}_{KD}$ has Lipschitz Hessian, then:
\begin{align*}
\nabla_{\boldsymbol{\widetilde{\theta}}}\mathcal{L}_{KD}(\mathcal{T}_{i};\boldsymbol{\widetilde{\theta}})&=\nabla_{\boldsymbol{\theta}}\mathcal{L}_{KD}(\mathcal{T}_{i};\boldsymbol{\theta})+O(\alpha^2)\\
&-\alpha\nabla_{\boldsymbol{\theta}}^2\mathcal{L}_{KD}(\mathcal{T}_{i};\boldsymbol{\theta})\nabla_{\boldsymbol{\theta}}\mathcal{L}_{KD}(\mathcal{T}_{j};\boldsymbol{\theta}),
\end{align*}
where $\alpha$ is the step size of the inner loop.
\end{lemma}

\begin{theorem}\label{theorem1}
If $\mathcal{T}_{i}$ can be regarded as independent identically distributed samples from the distribution $\mathcal{P}_M$, then:
\begin{align*}
\nabla_{\boldsymbol{\theta}}\mathcal{L}_{KD}(\mathcal{T}_{i};\boldsymbol{\widetilde{\theta}}) &=\nabla_{\boldsymbol{\theta}}\mathcal{L}_{KD}(\mathcal{T}_{i};\boldsymbol{\theta})+O(\alpha^2)\\
&-\alpha\nabla\underbrace{(\nabla_{\boldsymbol{\theta}}\mathcal{L}_{KD}(\mathcal{T}_{i};\boldsymbol{\theta})\nabla_{\boldsymbol{\theta}}\mathcal{L}_{KD}(\mathcal{T}_{j};\boldsymbol{\theta}))}_{Gradient\,Alignment},
\end{align*}
\ie, the inner product between gradients of different tasks.
\end{theorem}

From the analysis above, we observe that the gradient of $\mathcal{T}_{i}$ produces the inner product with other tasks that might pose conflicts. This indicates, optimizing the meta-learner minimizes the expected loss over tasks $\mathcal{T}_{i}\sim\mathcal{P}_{M}$ (effectively the ERM), and maximizes the inner product between gradients of different tasks $\nabla_{\boldsymbol{\theta}}\mathcal{L}_{KD}(\mathcal{T}_{i};\boldsymbol{\theta})\nabla_{\boldsymbol{\theta}}\mathcal{L}_{KD}(\mathcal{T}_{j};\boldsymbol{\theta})$. Hence, it enforces the meta-learner to seek a common direction, encouraging across-task generalization.

\begin{algorithm}[tb]
\small
\DontPrintSemicolon
\SetKwInOut{Input}{Input}\SetKwInOut{Output}{Output}\SetKwInOut{Require}{Require}
\textbf{Input: }{Meta-learner $F(\cdot;\boldsymbol{\theta})$; recovered tasks $\{\mathcal{T}_{i}\}_{i=1}^b$ and corresponding pre-trained models $\{M_{i}\}_{i=1}^b$ in a batch.
}\\
\textbf{Output: }{Optimized meta-learner $F(\cdot;\boldsymbol{\theta})$.}\\
Clone meta-learner $F(\cdot;\boldsymbol{\widetilde{\theta}})\leftarrow F(\cdot;\boldsymbol{\theta})$\\
\tcp{Multi-task knowledge distillation}
\For{ {\rm index} $i \leftarrow 1$ \KwTo $b$}{
Compute $\boldsymbol{g}_i= \nabla_{\boldsymbol{\widetilde{\theta}}}\mathcal{L}_{KD}(\mathcal{T}_{i};\boldsymbol{\widetilde{\theta}})$ \wrt \cref{eq:3}\\
$\text{Update }\boldsymbol{\widetilde{\theta}}\leftarrow\boldsymbol{\widetilde{\theta}}-\alpha\boldsymbol{g}_i$\\
}
\tcp{Outer loop for meta-learner}
$\text{Update }\boldsymbol{\theta}\leftarrow\boldsymbol{\theta}+\epsilon(\boldsymbol{\widetilde{\theta}}-\boldsymbol{\theta})$
\caption{ \textsc{BelL}}
\label{alg:alg2}
\end{algorithm}

\noindent
\textbf{Optimization.}
However, optimizing the meta-learner based on \cref{eq:3} requires high-order gradients when computing $\nabla_{\boldsymbol{\theta}}\mathcal{L}_{KD}(\mathcal{T}_{i};\boldsymbol{\widetilde{\theta}})$, making the back-propagation highly inefficient. We could apply a first-order approximation~\cite{nichol2018first} to further accelerate the gradient computing. We continue to update $\boldsymbol{\widetilde{\theta}}$ using the gradient $\nabla_{\boldsymbol{\widetilde{\theta}}}\mathcal{L}_{KD}(\mathcal{T}_{i};\boldsymbol{\widetilde{\theta}})$. Then, the meta-learner $F(\cdot;\boldsymbol{\theta})$ is updated through a weighted difference between the resulted parameters $\boldsymbol{\widetilde{\theta}}$ and the original parameters $\boldsymbol{\theta}$ (\cf \cref{alg:alg2}).

\noindent
\textbf{Cross Task Replay.}
A small number of pre-trained models (\eg, 100) are insufficient to represent the actual underlying task distribution, leading to an over-reliance on specific tasks. Hence, we employ a memory bank $\mathcal{B}$ with a first-in-first-out structure to store the previously recovered tasks, and replay new tasks interpolated across different tasks, \ie, a random combination of classes. This further alleviates potential conflicts between tasks, assisting the meta-learner in generalizing across tasks. For these replayed tasks $\hat{\mathcal{T}}$, which include the support set $\boldsymbol{S}$, the query set $\boldsymbol{Q}$ and class labels $\boldsymbol{Y}$, we update the meta-learner as follows:
\begin{align}
\label{eq:6}
\min_{\boldsymbol{\theta}}\mathbb{E}_{\hat{\mathcal{T}}\sim \mathcal{P}_{B}}\mathcal{L}_{outer}
\triangleq&\mathcal{L}_{CE}(F(\boldsymbol{Q};\boldsymbol{\theta}_c),\boldsymbol{Y}_{\boldsymbol{Q}}),\\
\mathrm{s.t.~}\boldsymbol{\theta}_{c}
=\min_{\boldsymbol{\theta}}\mathcal{L}_{inner}&\triangleq\mathcal{L}_{CE}(F(\boldsymbol{S};\boldsymbol{\theta}),\boldsymbol{Y}_{\boldsymbol{S}}). \nonumber
\end{align}

In the end, we summarize the main pipeline of the proposed framework in \cref{alg:alg3}.

\begin{algorithm}[tb]
\small
\DontPrintSemicolon
\SetKwInOut{Input}{Input}\SetKwInOut{Output}{Output}\SetKwInOut{Require}{Require}
\textbf{Input: }{A collection of pre-trained models $\mathcal{M}_{pool}$;\,\,\,\,\,\,\,\,\,\,\,\,\,\,\,\,\,\, a meta-generator $G(\cdot;\boldsymbol{\theta}_G)$; the memory bank $\mathcal{B}$.
}\\
\textbf{Output: }{A meta-learner $F(\cdot;\boldsymbol{\theta})$ for new unseen tasks.}\\

\For{ {\rm epoch} $i \leftarrow 1$ \KwTo $N$}{
Sample pre-trained models $\{M_{i}\}_{i=1}^b$ from $\mathcal{M}_{pool}$\\
Recover specific tasks $\{\mathcal{T}_{i}\}_{i=1}^b$ with $G(\cdot;\boldsymbol{\theta}_G)$\\
Put into memory bank $\mathcal{B} \leftarrow \mathcal{B} + \{\mathcal{T}_{i}\}_{i=1}^b$\\
Update meta-generator $\boldsymbol{\theta}_G$ \wrt \cref{eq:2}\\
Update meta-learner $\boldsymbol{\theta}$ with $\{M_{i}\}_{i=1}^b$ and $\{\mathcal{T}_{i}\}_{i=1}^b$\\
Construct interpolated tasks $\hat{\mathcal{T}}$ from $\mathcal{B}$\\
Update meta-learner $\boldsymbol{\theta}$ \wrt \cref{eq:6}\\
}

\caption{ \textsc{Free}}
\label{alg:alg3}
\end{algorithm}

\section{Experiments}
In this section, we begin by detailing our experimental setup, including an overview of the compared baselines. Following that, we present the main results of our approach, evaluating performance based on meta-testing accuracy and meta-training speed. Additionally, we offer ablation studies and discussions to facilitate comprehensive analyses.

\subsection{Experimental Setup}

\noindent
\textbf{Datasets and pre-trained models.}
We conduct experiments on two widely-used DFML benchmark datasets, and one fine-grained dataset, including \emph{miniImageNet}~\cite{miniimagenet}, \emph{CIFAR-FS}~\cite{cifarfs} and \emph{CUB}~\cite{WahCUB200_2011}. Following standard splits~\cite{wertheimer2021few}, we split each dataset into the meta-training, meta-validating and meta-testing subsets with disjoint label spaces.
In the DFML setting, we have no access to the meta-training data. Following \cite{wang2022meta,hu2023learning}, we collect 100 models pre-trained on 100 $N$-way tasks sampled from the meta-training subset, and those models are used as the meta-training resources.

\noindent
\textbf{Implementation details.}
For the model architecture, we adopt Conv4 as the architecture of the meta-learner and the pre-trained models for a fair comparison with existing works. We provide the detailed structure for the meta-generator in Appendix B. For hyperparameters, the batch size $b$ is set to 4, and the learning rate $\gamma$, $\alpha$, and $\epsilon$ are all set to 0.001.
We report the average accuracy over 600 meta-testing tasks. We leave the other setup in Appendix A.

\noindent
\textbf{Baselines.}
\textbf{(i) Random.} Learn a classifier using the support set from scratch for each meta-testing task. \textbf{(ii) Best-Model.} We select the pre-trained model with the highest reported accuracy to directly predict the query set during meta-testing. \textbf{(iii) Average.} Average all pre-trained models and then finetune it using the support set. \textbf{(iv) OTA}~\cite{singh2020model}. Calculate the weighted average of all pre-trained models and then finetune it using the support set. \textbf{(v) DRO}~\cite{wang2022meta}. Meta-learn a hyper-network to fuse all pre-trained models into one single model, which serves as the meta-initialization and can be adapted to each meta-testing task using the support set. \textbf{(vi) PURER} \citep{hu2023architecture}. Adversarially train the meta-learner with a learnable dataset, where a batch of pseudo tasks is sampled for meta-training at each iteration. \textbf{(vii) BiDf-MKD}~\cite{hu2023learning}. A bi-level data-free meta knowledge distillation framework to transfer general knowledge in the white-box setting.

\begin{table*}[tb]
\footnotesize
\centering
\renewcommand\arraystretch{1.1}
\setlength{\tabcolsep}{4.85pt}
\caption{\small Compare to existing baselines in DFML. Time: GPU hours taken by the data recovery process, so non-inversion methods only report accuracy. $\textsc{Free}_{2/5}$ denotes the 2-step and 5-step adaptation of the meta-generator, respectively.}
\vspace{-1em}
\begin{tabular}{lccccccccc}
			\addlinespace
			\toprule
			\specialrule{0em}{1pt}{1pt}
			\multirow{2}{*}{ \bf Method} &\multicolumn{3}{c}{\emph {CIFAR-FS}} & \multicolumn{3}{c}{\emph {miniImageNet}} & \multicolumn{3}{c}{\emph {CUB}}\\ 
			\cmidrule(l){2-4} \cmidrule(l){5-7} \cmidrule(l){8-10}
			\specialrule{0em}{1pt}{1pt}
			&{$5$-way $1$-shot} & {$5$-way $5$-shot}& {Time} &{$5$-way $1$-shot} &{$5$-way $5$-shot} & {Time} &{$5$-way $1$-shot} &{$5$-way $5$-shot}& {Time}\\
			\midrule
			\specialrule{0em}{1pt}{1pt}
			Random  & {28.59} {\scriptsize $\pm$ 0.56} & {34.77} {\scriptsize $\pm$ 0.62} & {-} &{25.06} {\scriptsize $\pm$ 0.50}& {28.10} {\scriptsize $\pm$ 0.52} & {-} &{26.26} {\scriptsize $\pm$ 0.48}& {29.89} {\scriptsize $\pm$ 0.55}& {-}\\
			\specialrule{0em}{1pt}{1pt}
           Best-Model  & {21.68} {\scriptsize $\pm$ 0.66} & {25.05} {\scriptsize $\pm$ 0.67}& {-} & {22.86} {\scriptsize $\pm$ 0.61}& {26.26} {\scriptsize $\pm$ 0.63}& {-} & {24.16} {\scriptsize $\pm$ 0.73}& {29.16} {\scriptsize $\pm$ 0.73}& {-}\\
           \specialrule{0em}{1pt}{1pt}
           Average  & {23.96} {\scriptsize $\pm$ 0.53} & {27.04} {\scriptsize $\pm$ 0.51}& {-} & {23.79} {\scriptsize $\pm$ 0.48}& {27.49} {\scriptsize $\pm$ 0.50}& {-} & {24.53} {\scriptsize $\pm$ 0.46}& {28.00} {\scriptsize $\pm$ 0.47}& {-}\\
           \specialrule{0em}{1pt}{1pt}
			OTA~\cite{singh2020model} &  {29.10} {\scriptsize $\pm$ 0.65}  & {34.33} {\scriptsize $\pm$ 0.67}& {-} & {24.22}  {\scriptsize $\pm$ 0.53} & {27.22} {\scriptsize $\pm$ 0.59}& {-} & {24.23} {\scriptsize $\pm$ 0.60}& {25.42} {\scriptsize $\pm$ 0.63}& {-} \\
    		\specialrule{0em}{1pt}{1pt}
			DRO~\cite{wang2022meta}  & {30.43} {\scriptsize $\pm$ 0.43} & {36.21} {\scriptsize $\pm$ 0.51}& {-} & {27.56} {\scriptsize $\pm$ 0.48} & {30.19} {\scriptsize $\pm$ 0.43}& {-} & {28.33} {\scriptsize $\pm$ 0.69} & {31.24} {\scriptsize $\pm$ 0.76}& {-}\\
            \specialrule{0em}{1pt}{1pt}
            \midrule
            PURER~\cite{hu2023architecture}  & {38.66} {\scriptsize $\pm$ 0.78} & {51.95} {\scriptsize $\pm$ 0.79}& {1.21h} & {31.14} {\scriptsize $\pm$ 0.63} &  {40.86} {\scriptsize $\pm$ 0.64}& {1.31h}& {30.08} {\scriptsize $\pm$ 0.59} &  {40.93} {\scriptsize $\pm$ 0.66}& {1.97h}\\
			\specialrule{0em}{1pt}{1pt}
            BiDf-MKD~\cite{hu2023learning}  & {37.66} {\scriptsize $\pm$ 0.75} & {51.16} {\scriptsize $\pm$ 0.79}& {2.47h} & {30.66} {\scriptsize $\pm$ 0.59} & {42.30} {\scriptsize $\pm$ 0.64}& {8.87h} & {31.62} {\scriptsize $\pm$ 0.60} & {44.32} {\scriptsize $\pm$ 0.69}& {7.16h}\\
            \specialrule{0em}{1pt}{1pt}
            \midrule
            \textbf{\textsc{Free}$_2$}  &   {36.56 {\scriptsize $\pm$ 0.73}} & {47.31 {\scriptsize  $\pm$ 0.76}}& {0.05h} & {30.06 {\scriptsize  $\pm$ 0.61}} & {41.60 {\scriptsize  $\pm$ 0.68}}& {0.25h} &   {30.64 {\scriptsize $\pm$ 0.58}} & {39.43 {\scriptsize  $\pm$  0.61}}& {0.20h}\\ 
            \specialrule{0em}{1pt}{1pt}
           \textbf{\textsc{Free}$_5$}   &   \bf {39.13 {\scriptsize $\pm$ 0.85}} & \bf {52.58 {\scriptsize  $\pm$ 0.77}}& {0.11h} & \bf {33.03 {\scriptsize  $\pm$ 0.69}} & \bf {45.45 {\scriptsize  $\pm$ 0.69}}& {0.41h}&   \bf {31.94 {\scriptsize $\pm$ 0.61}} & \bf {49.10 {\scriptsize  $\pm$ 0.68}}& {0.32h}\\ 
           \bottomrule
	\end{tabular}
\vspace{-1.5em}
\label{table:mainres1}
\end{table*}

\begin{table}[tb]
\footnotesize
\centering
\renewcommand\arraystretch{1.1}
\setlength{\tabcolsep}{4.85pt}
\caption{\small Compare to baselines in a multi-domain scenario.}
\vspace{-1em}
\begin{tabular}{lccc}
			\addlinespace
			\toprule
			\specialrule{0em}{1pt}{1pt}
			\multirow{2}{*}{ \bf Method} &\multicolumn{3}{c}{\emph {CIFAR-FS} + \emph {miniImageNet} + \emph {CUB}}\\ 
			\cmidrule(l){2-4} 
			\specialrule{0em}{1pt}{1pt}
			&{$5$-way $1$-shot} & {$5$-way $5$-shot}& {Time} \\
			\midrule
			\specialrule{0em}{1pt}{1pt}
			Random  & {24.85} {\scriptsize $\pm$ 0.54} & {28.35} {\scriptsize $\pm$ 0.61} & {-}\\
			\specialrule{0em}{1pt}{1pt}
            PURER~\cite{hu2023architecture}  & {29.67} {\scriptsize $\pm$ 0.59} & {37.96} {\scriptsize $\pm$ 0.60}& {3.43h} \\
			\specialrule{0em}{1pt}{1pt}
            BiDf-MKD~\cite{hu2023learning}  & {31.44} {\scriptsize $\pm$ 0.64} & {40.96} {\scriptsize $\pm$ 0.59}& {7.78h} \\
            \specialrule{0em}{1pt}{1pt}
            \midrule
            \textbf{\textsc{Free}$_2$}  &   {30.39 {\scriptsize $\pm$ 0.56}} & {40.69 {\scriptsize  $\pm$ 0.68}}& {0.20h} \\ 
            \specialrule{0em}{1pt}{1pt}
           \textbf{\textsc{Free}$_5$}   &   \bf {31.51 {\scriptsize $\pm$ 0.63}} & \bf {44.04 {\scriptsize  $\pm$ 0.66}}& {0.34h}\\ 
           \bottomrule
	\end{tabular}
\vspace{-1em}
\label{table:mainres2}
\end{table}

\noindent
\textbf{Metrics.}
In addition to the standard few-shot classification accuracy, we also focus on the speed of model inversion in DFML, quantifying this through the metric of GPU hours. Note that the time cost of meta-learner training is omitted since we only focus on the data recovery process. For consistency in comparison, all GPU hours are obtained using a single GeForce RTX 3090 GPU.

\subsection{Main Results}

\noindent
\textbf{Comparisons with baselines.}
\cref{table:mainres1} shows the results for 5-way classification compared with current baselines. To ascertain the efficacy of \textsc{Free}, we compare it against two DFML algorithm categories: the non-inversion methods, which facilitate model fusion in the parameter space, and the inversion-based methods that generate pseudo data for training. As shown in \cref{table:mainres1}, our approach considerably outperforms all non-inversion methods, and is significantly faster than other inversion-based methods. For instance, PURER generates data by constructing a smaller pseudo dataset, with each class comprising only 20 images. Nevertheless, PURER demands 15000 iterations to converge in optimizing the dataset. In contrast, we only require 2/5 steps for different tasks, owning to our adaptive meta-generator.

Beyond speed enhancements, \textsc{Free} also achieves better accuracy across all three datasets. Compared with the foremost baseline BiDf-MKD, \textsc{Free} displays performance enhancements ranging from 0.32\%$\sim$2.37\% for 1-shot learning and from 1.42\%$\sim$4.78\% for 5-shot learning. These improvements can be attributed to two key factors. Firstly, our proposed \textsc{BelL} effectively aligns conflicting gradients across different tasks, directly improving the generalization capabilities. Secondly, the efficacy of our meta-generator is evident as it rapidly absorbs knowledge from pre-trained models and provides valuable training data.

\noindent
\textbf{Multi-domain scenario.}
We conduct experiments in a challenging multi-domain scenario where all pre-trained models are tailored to address distinct tasks from multiple meta-training subsets (\emph{CIFAR-FS}, \emph{miniImageNet}, and \emph{CUB}). For meta-testing, we assess the meta-learner on unseen tasks spanning \emph{CIFAR-FS}, \emph{miniImageNet}, and \emph{CUB} concurrently, demanding the meta-learner to possess generalization capabilities across multiple domains. \cref{table:mainres2} showcases our results. Our approach achieves commendable results, outperforming the baseline by 6.66\% and 15.69\% in 1-shot and 5-shot learning, respectively. This solidly demonstrates that \textsc{BelL} effectively aligns tasks from different domains, optimizing them towards a unified direction, resulting in robust generalization across various domains.

\begin{table}[tb]
\footnotesize
\centering
\renewcommand\arraystretch{1.1}
\setlength{\tabcolsep}{4.85pt}
\caption{\small Compare to baselines in a multi-architecture scenario.}
\vspace{-1em}
\begin{tabular}{lccc}
			\addlinespace
			\toprule
			\specialrule{0em}{1pt}{1pt}
			\multirow{2}{*}{ \bf Method} &\multicolumn{3}{c}{\emph {CIFAR-FS}}\\ 
			\cmidrule(l){2-4} 
			\specialrule{0em}{1pt}{1pt}
			&{$5$-way $1$-shot} & {$5$-way $5$-shot}& {Time} \\
			\midrule
            PURER~\cite{hu2023architecture}  & {39.15} {\scriptsize $\pm$ 0.70} & {49.08} {\scriptsize $\pm$ 0.74}& {1.75h} \\
			\specialrule{0em}{1pt}{1pt}
            BiDf-MKD~\cite{hu2023learning}  & {38.08} {\scriptsize $\pm$  0.80} & {50.58} {\scriptsize $\pm$ 0.81}& {4.33h} \\
            \specialrule{0em}{1pt}{1pt}
            \midrule
           \textbf{\textsc{Free}$_5$}   &   \bf {39.63 {\scriptsize $\pm$ 0.82}} & \bf {52.12 {\scriptsize  $\pm$ 0.79}}& {0.13h}\\ 
           \bottomrule
	\end{tabular}
\vspace{-0.5em}
\label{table:mainres3}
\end{table}

\noindent
\textbf{Multi-architecture scenario.}
We also conduct experiments in a multi-architecture scenario where each pre-trained model differs in architecture. For each pre-trained model, the architecture is randomly chosen from Conv4, ResNet-10 and ResNet-18. The results, presented in \cref{table:mainres3}, demonstrate the efficacy of our approach. It outperforms all baselines and can apply to multi-architecture scenario without any change. This flexibility is attributable to our method's lack of constraints on the underlying architecture or scale of the pre-trained models, which enables effective task alignment across heterogeneous models.
\subsection{Discussions}

\begin{table}[tb]
  \footnotesize
    \centering
    \renewcommand\arraystretch{1.1}
    \setlength{\tabcolsep}{4.85pt}
  \caption{Effects of the proposed modules.}
    \vspace{-1em}
  \begin{tabular}{lcccc}
    \toprule
    \multirow{2}{*}{\textbf{Setting}} & \multicolumn{2}{c}{Module} & \multicolumn{2}{c}{\emph {CIFAR-FS}} \\
    \cmidrule(r){2-3}
    \cmidrule(r){4-5}
    &\textsc{FIve}&\textsc{BelL}&5-way 1-shot&5-way 5-shot\\
    \midrule
    Baseline&&&36.33 {\scriptsize $\pm$ 0.81}&47.67 {\scriptsize $\pm$ 0.75}\\
    ERM&\checkmark&& 37.43 {\scriptsize $\pm$ 0.77} &49.54 {\scriptsize $\pm$ 0.73} \\
    Sequence&&\checkmark&38.56 {\scriptsize $\pm$ 0.78}&50.78 {\scriptsize $\pm$ 0.79}\\
    \midrule
    \textbf{\textsc{Free}$_5$}&\checkmark&\checkmark&\bf {39.13 {\scriptsize $\pm$ 0.85}}&\bf {52.58 {\scriptsize $\pm$ 0.77}}\\
     \bottomrule
  \end{tabular}
  \label{tab:ablation}
  \vspace{-1em}
\end{table}

\noindent
\textbf{Ablation studies.}
\cref{tab:ablation} evaluates the impact of each module on the \emph{CIFAR-FS}. Initially, we set a baseline only performing meta-learning via cross task replay, using a sequential generator. The generator trained on a previous pre-trained model is directly used as an initialization for subsequent tasks. However, updating the generator with only 5 steps for each pre-trained model proves to be insufficient. This is attributed to the non-reusability of the sequential generator across diverse pre-trained models.
Incorporating \textsc{FIve} module offers notable enhancements, suggesting that the meta-generator can rapidly adapt and inverse task-specific training data. Nonetheless, the current method (\ie, ERM) still merely aggregates knowledge distillation losses from multiple tasks to update the meta-learner, overlooking the alignment of optimization directions across tasks. By adding \textsc{BelL} module, we observe a significant improvement (\ie, Sequence), demonstrating the effectiveness of the conflicting gradient alignment. With all modules, we achieve the best performance with a boosting improvement of 2.8\% and 4.91\%, showing their complementarity.

\begin{figure}[tb]
  \centering
  \begin{minipage}[t!]{0.48\linewidth}
    \includegraphics[width=\linewidth]{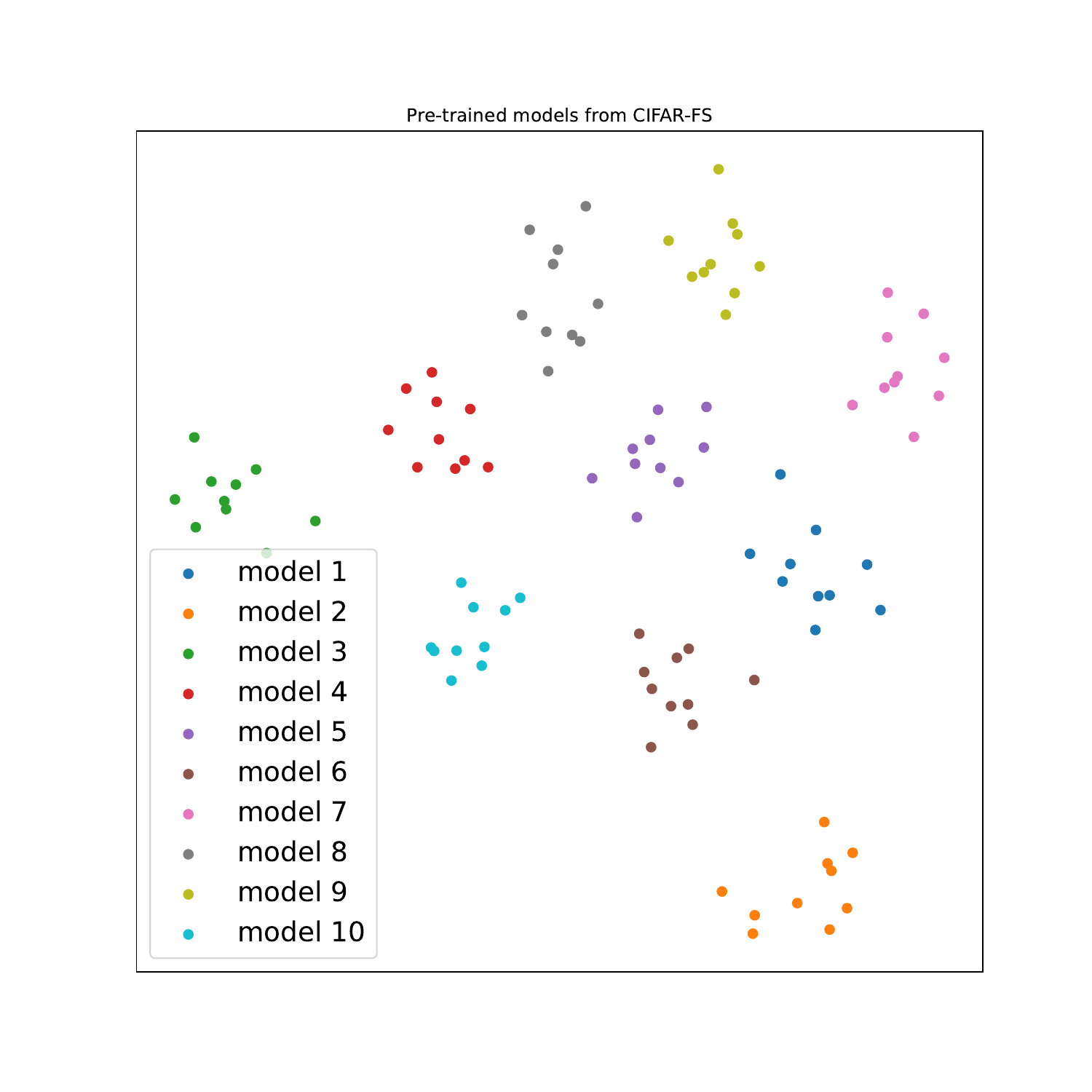}
    \caption{The $t$-SNE visualization of same samples input into different pre-trained models.}
    \label{fig:domain}
  \end{minipage}
  \hfill
  \begin{minipage}[t!]{0.49\linewidth}
    \includegraphics[width=\linewidth]{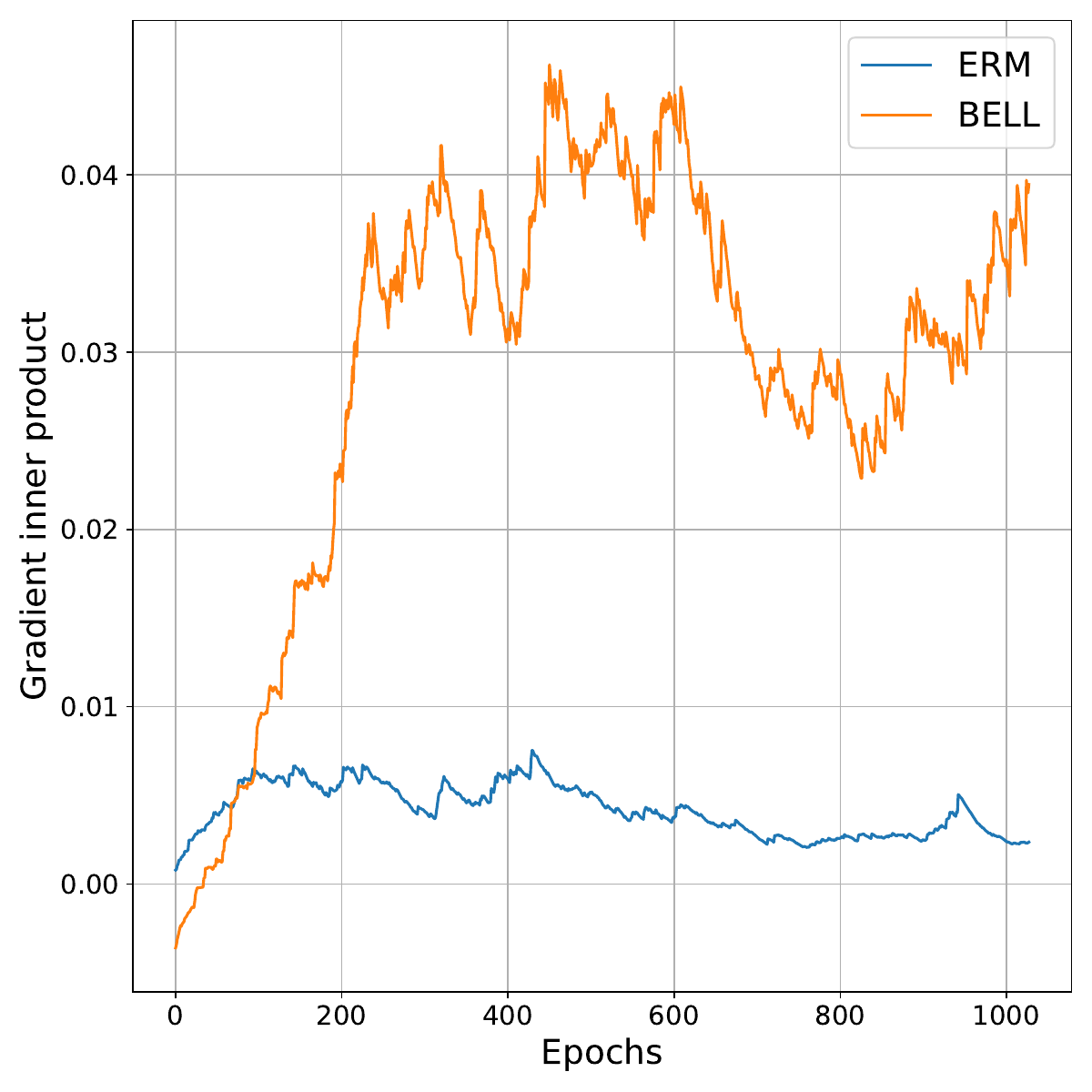}
    \caption{The gradient inner product across tasks from different pre-trained models.}
    \label{fig:gip}
  \end{minipage}
  \vspace{-1em}
\end{figure}

\noindent
\textbf{Heterogeneity among pre-trained models.}
We present a $t$-SNE visualization of the features extracted from different pre-trained models in \cref{fig:domain}. We randomly sample 10 images of the same class and use 10 pre-trained models from \emph{CIFAR-FS} to extract their features. The visualization clearly shows that the features of the same image, when extracted by different pre-trained models, are scattered in the feature space. This implies that the inversed tasks from distinct pre-trained models naturally inherit varying feature distributions. This observation underscores the importance of addressing the heterogeneity among pre-trained models.

\noindent
\textbf{Gradient inner product maximization.}\label{sec:bell}
In our approach, we incorporate \textsc{BelL} to align gradients across different tasks, guiding the meta-learner towards a unified direction. To further validate whether \textsc{BelL} implicitly maximizes the gradient inner product, we chart its evolution during meta-learner training in \cref{fig:gip}. We contrast \textsc{BelL} with the ERM method, monitoring the mean gradient inner product between tasks derived from varied pre-trained models in each inner loop. As evident from \cref{fig:gip}, the gradient inner product for \textsc{BelL} gradually rises with training progression. Conversely, ERM maintains lower values throughout. This confirms \textsc{BelL}'s effectiveness in implicitly aligning the gradient direction by enhancing the gradient inner product.

\begin{figure}[t]
\begin{minipage}[t]{\linewidth}
        \centering
        \subfloat[5-way 1-shot]{\includegraphics[width=0.49\columnwidth]{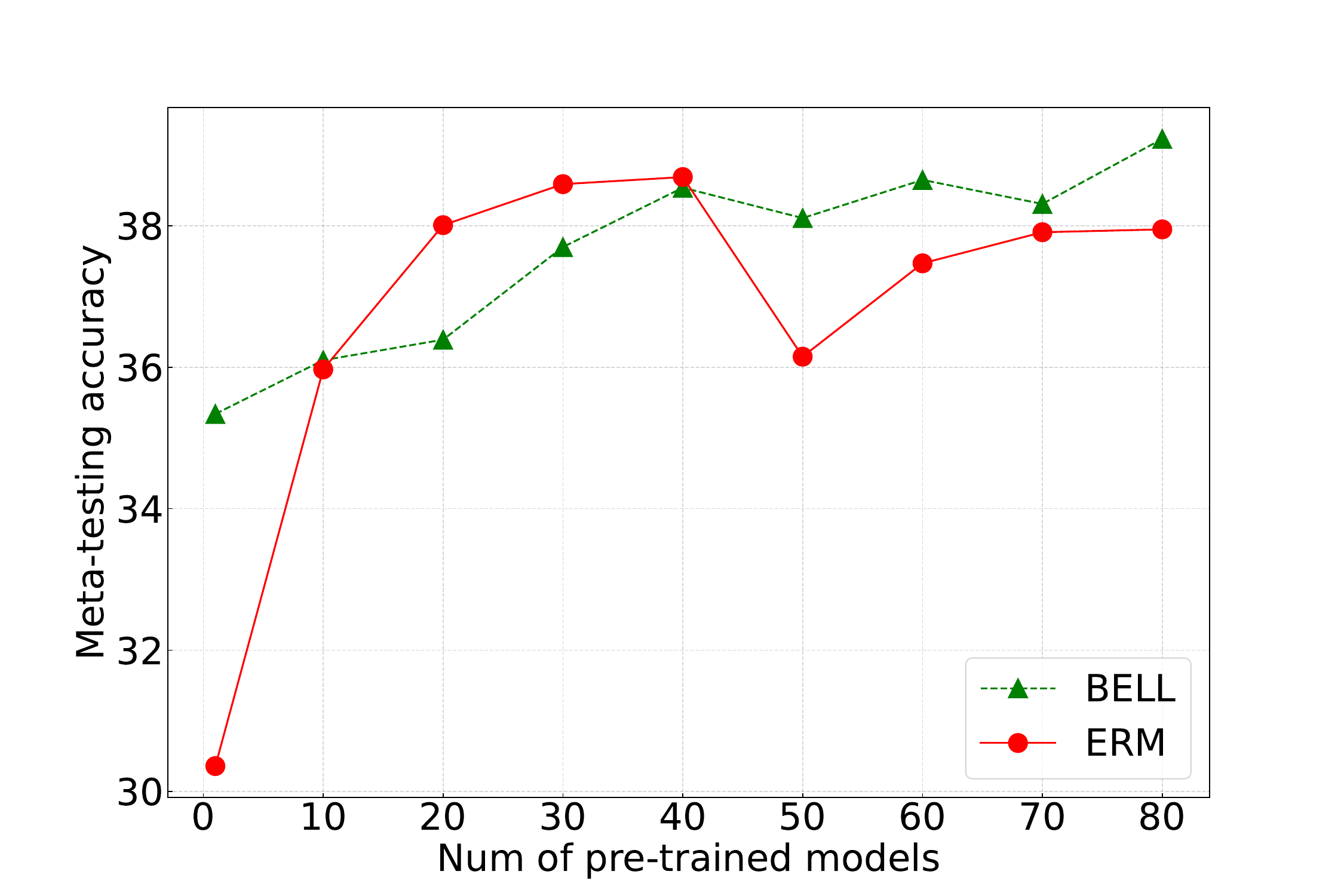} }
        \subfloat[5-way 5-shot]{\includegraphics[width=0.49\columnwidth]{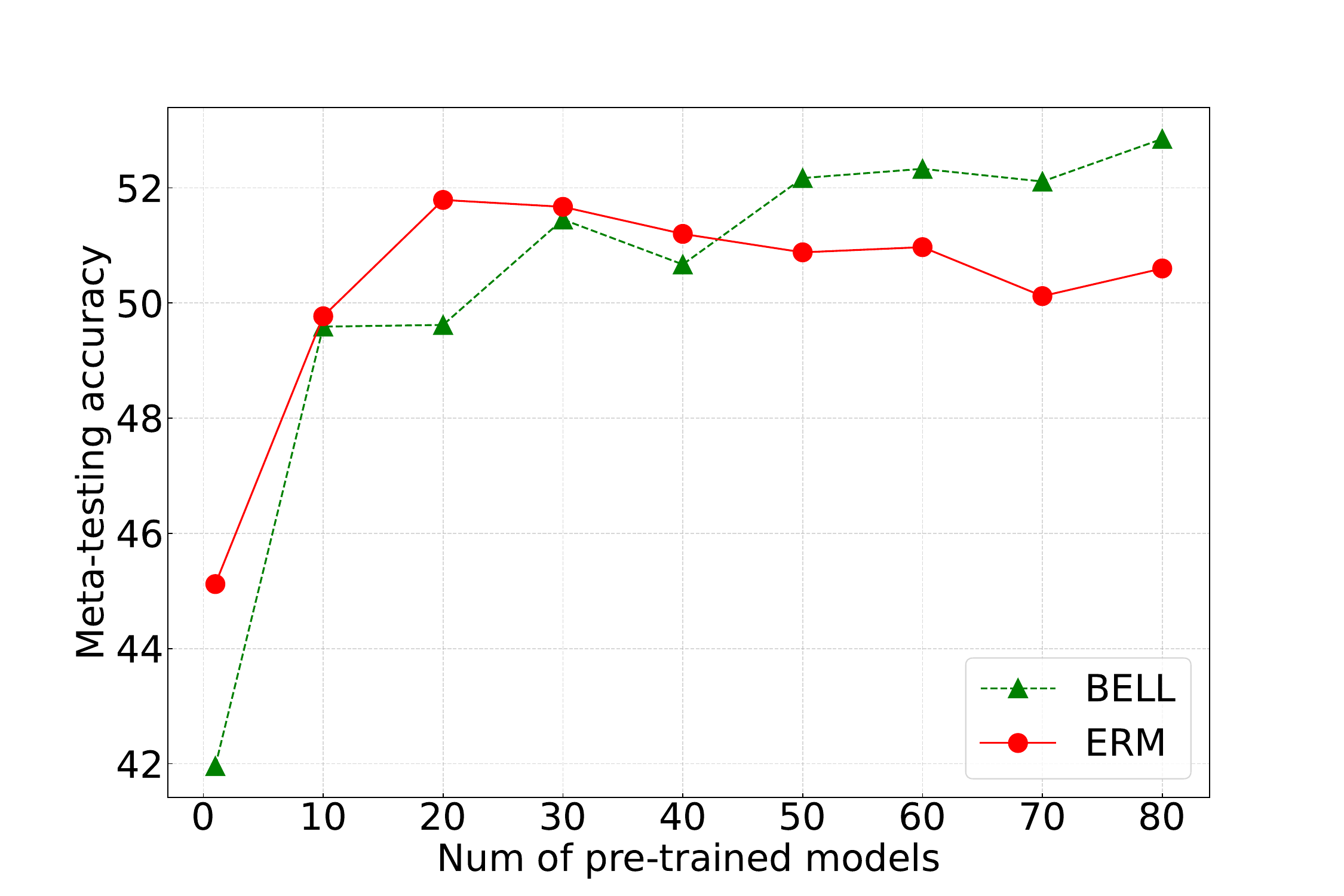} }
        \vspace{-1em}
         \caption{\small Effects of the number of pre-trained models.}
	\label{fig:number}
\end{minipage}
\vspace{-0.5em}
\end{figure}

\noindent
\textbf{Effects of the number of pre-trained models.}
We conduct experiments on DFML with varying numbers of pre-trained models. Each model is pre-trained on a 5-way subset of \emph{CIFAR-FS}. As shown in \cref{fig:number}, the performance improves with an increasing number of models. This can be attributed to the fact that a greater quantity of pre-trained models offers a more comprehensive understanding of various classes, thereby enhancing generalization capabilities for unseen classes. Interestingly, we observe that ERM delivers promising results with a smaller set of pre-trained models, at times even outperforming our approach. However, as the number of pre-trained models continues to climb, ERM's performance starts to drop and is eventually surpassed by our approach. This decline can be attributed to the growing issues of model heterogeneity and task conflicts with an increase in the number of pre-trained models. The poor performance of ERM highlights the challenge of utilizing numerous available pre-trained models from sources like GitHub to enable efficient learning on unseen downstream tasks. Contrarily, we effectively navigate this challenge, consistently delivering performance enhancements.

\begin{figure}[t]
  \centering
    \includegraphics[width=0.9\linewidth]{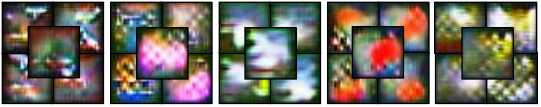}
    \vspace{-0.5em}
   \caption{Visualizations of the 5-way recovered task via our meta-generator. Each group represents one class.}
   \label{fig:visualization}
   \vspace{-1em}
\end{figure}

\noindent
\textbf{Visualizations.}
\cref{fig:visualization} presents a 5-way task recovered from the Conv4. A significant advantage of DFML lies in its ability to learn from weaker pre-trained models without data privacy leakage. Our approach generates images that are visually distinct from the originals, thereby addressing privacy concerns. The images we generate display high-frequency patterns and textures, which are recognized for their particular value in training neural networks~\cite{geirhos2018imagenet,micaelli2019zero}.
\section{Conclusion}
In this work, we highlight the necessity of addressing the efficiency dilemma and the heterogeneity among pre-trained models in DFML. To address the slow recovery speed, we train a meta-generator capable of rapidly adapting to specific tasks. To further address the potential conflicts in recovered tasks, we train a meta-learner to align the gradients of different tasks. This meta-learner captures task-invariant features, thus enabling generalization to new unseen tasks. Extensive experiments on multiple benchmarks show significant speed and performance gains in our approach.

\vspace{0.15cm}

\noindent
\textbf{Acknowledgement.}\ This work is supported by the National Key R\&D Program of China (2022YFB4701400/4701402), SSTIC Grant(KJZD20230923115106012), Shenzhen Key Laboratory (ZDSYS20210623092001004), and Beijing Key Lab of Networked Multimedia.

\clearpage
{
    \small
    \bibliographystyle{ieeenat_fullname}
    \bibliography{main}
}

\end{document}